%% file: main_IVC.tex
\documentclass[final,5p]{elsarticle}

\input{packages}

\title{Open-Set Face Recognition with Maximal Entropy and Objectosphere Loss}
\author[ufmg,uzh]{Rafael Henrique Vareto}
\author[uzh]{Yu Linghu}
\author[uccs]{Terrance Edward Boult}
\author[ufmg]{William Robson Schwartz}
\author[uzh,uccs]{Manuel G\"unther}

\address[ufmg]{Smart Sense Laboratory, Federal University of Minas Gerais, Belo Horizonte, Brazil -- \{rafaelvareto,william\}@dcc.ufmg.br}
\address[uzh]{Department of Informatics, University of Z\"urich, Z\"urich, Switzerland -- \{linghu,guenther\}@ifi.uzh.ch}
\address[uccs]{Vision and Security Technology Lab, University of Colorado Colorado Springs, Colorado Springs, USA -- tboult@vast.uccs.edu}

\usepackage{fancyhdr}

\begin{document}


    \input{sections/abstract}
    \maketitle

    \thispagestyle{empty}

    {
    \chead{\footnotesize This is a pre-print of the original paper to be published in Image and Vision Computing 2023.}
    \lhead{}
    \thispagestyle{fancy}
    }

    \input{sections/introduction}
    \input{sections/related}
    \input{sections/approach}
    \input{sections/experiments}

\input{sections/discussion}
    \input{sections/conclusion}

    \biboptions{sort&compress}

    \bibliographystyle{unsrt}
    \bibliography{bibliography}

\end{document}

%% file: packages.tex
\usepackage{amsmath,amssymb,amsfonts}
\usepackage{algorithmic}
\usepackage{graphicx}
\usepackage{textcomp}
    
\usepackage{color,soul}
\usepackage{comment}
\usepackage[inline]{enumitem}
\usepackage[pagebackref=true,breaklinks=true,colorlinks,bookmarks=false]{hyperref}
\usepackage{lineno}
\usepackage{lipsum}
\usepackage{listings}
\usepackage{multirow}
\usepackage[listofformat=parens, justification=centering, subrefformat=subparens]{subfig}

\usepackage[usenames,dvipsnames]{xcolor}

\setlist[enumerate]{nosep}

\graphicspath{{graphics/authors/}{graphics/figures/}{graphics/images/}}

\newcommand\ie{\textit{i.\,e.}}
\DeclareMathOperator*{\argmax}{\arg\max}
\newcommand\capt[3]{\caption[#2]{\label{#1}\fontsize{9}{11}\textsc{#2}. \fontsize{8}{10}\textit{#3}}}
\newcommand\etal[1]{\textit{et al.}~\cite{#1}}
\newcommand\fig[1]{Fig.~\ref{fig:#1}}
\newcommand\sfig[1]{Fig.~\subref{fig:#1}} 
\newcommand\tab[1]{Tab.~\ref{tab:#1}}
\newcommand\mcbf[3]{\multicolumn{#1}{#2}{\textbf{#3}}}

\newcommand{\rone}[1]{{#1}}
\newcommand{\rtwo}[1]{{#1}}
\newcommand{\rthree}[1]{{#1}}



%% file: sections/abstract.tex
\begin{abstract}
    Open-set face recognition characterizes a scenario where unknown individuals, unseen during the training and enrollment stages, appear on operation time.
    This work concentrates on watchlists, an open-set task that is expected to operate at a low False Positive Identification Rate and generally includes only a few enrollment samples per identity.
    We introduce a compact adapter network that benefits from additional \emph{negative} face images when combined with distinct cost functions, such as Objectosphere Loss (OS) and the proposed Maximal Entropy Loss (MEL).
    MEL modifies the traditional Cross-Entropy loss in favor of increasing the entropy for negative samples and attaches a penalty to known target classes in pursuance of gallery specialization.
    The proposed approach adopts pre-trained deep neural networks (DNNs) for face recognition as feature extractors.
    Then, the adapter network takes deep feature representations and acts as a substitute for the output layer of the pre-trained DNN in exchange for an agile domain adaptation.
    Promising results have been achieved following open-set protocols for three different datasets: LFW, IJB-C, and UCCS as well as state-of-the-art performance when supplementary negative data is properly selected to fine-tune the adapter network.\\
\end{abstract}


%% file: sections/introduction.tex
\section{Introduction}\label{sec:introduction}
In open-set face recognition, there is no guarantee that a person caught on camera has been previously enrolled in the gallery of known individuals.
Within the open-set task, there are \textit{watchlists}, a scenario that must operate at a very low False Positive Identification Rate (FPIR) as a foresight that the majority of queried individuals are not expected to be registered in the gallery set.
When a detected face is mistakenly assigned to one of the identities, it raises a false alarm (a false positive identification) that usually triggers human actions and, therefore, must be avoided to decrease both operational cost and personal discomfort of innocent citizens~\cite{hill2020wrongfully,gunther2020watchlist}.
Additionally, subjects of interest may be either missed by a face detector or erroneously classified as unknown individuals or assigned a different identity.

Face biometric systems using deep convolutional neural networks have matured into an age of ubiquitous deployment and high performance in recent years.
However, most researchers have left open-set problems aside and channeled their efforts into closed-set identification and verification applications.
Recently, an outstanding vendor of face recognition technology suffered considerable criticism for matching USA congress members to mugshots of criminals~\cite{romm2017amazons}.
The incident became an eye-opener on the risks of such commercial identification systems as false alarms can substantially bias security personnel while increasing the responsibility for officers to thoroughly verify the results of the surveillance system.  
After all, no one would be contented with innocent people being held up by law-enforcement agencies due to a biometric system error.

\begin{figure*}[t!]
  \centering
  \subfloat[Known classes only (no negatives)\label{fig:teaser:cel}]{\includegraphics[height=.25\textwidth,page=3]{graphics/figures/mnist/MNIST-Softmax-2}}
  \hspace*{0.01\columnwidth}
  \subfloat[Extra class with negatives (Garbage)\label{fig:teaser:gbg}]{\includegraphics[height=.25\textwidth,page=3]{graphics/figures/mnist/MNIST-BGSoftMax-2}}
  \hspace*{0.01\columnwidth}
  \subfloat[Objectosphere Adaptation\label{fig:teaser:os}]{\includegraphics[height=.25\textwidth,page=3]{graphics/figures/mnist/MNIST-EOS-2}}
  \capt{fig:teaser}{Boosting unknown detection with negative samples}{
    The behavior of three different approaches when trained with additional negative data and evaluated with unknown samples. 
    LeNet++ network~\cite{wen2016centerloss} topologies are trained on 10 MNIST classes (knowns, colored dots) and evaluated with EMNIST letters (negatives, black) as well as Devanagari letters (unknowns, gray).
  }
\end{figure*}

Neural networks are biased toward the data they have been trained on and rarely work well with unknown classes.
\fig{teaser}, adapted from Dhamija~\etal{dhamija2018objectosphere}, illustrates such behavior on a handwritten digit and character recognition task.
Charts \subref*{fig:teaser:cel} and \subref*{fig:teaser:gbg} demonstrate that unknown samples (gray dots) cover most of the known classes when the cross-entropy loss is employed, which proved to be insufficient for open-set problems.
Contrarily, adopting a cost function that duly handles negative samples attains better class separation and achieves superior performance.

Although the illustration \subref{fig:teaser:os} may hold true for elementary problems holding abundant samples and very few classes, it is not guaranteed that such behavior would propagate to more demanding biometric applications~\cite{gunther2020watchlist,palechor2023protocols}.
In favor of investigating neglected real-world face problems, this study evaluates how open-set loss functions assist neural networks when the training data consists of a few instances per identity. 
We propose Maximal Entropy Loss (MEL), a function that adds a penalty margin to known identities and increases the entropy for negative samples as it guides a network into differentiating unknown from known subjects.
\rone{We also implement with an adapter network that is quickly trained by inputting deep features obtained with leading face architectures and avoid retraining deep backbones every time the gallery set is updated.}

This work discloses how a compact adaptation network, equipped with few fully-connected layers, responds to open-set protocols on three different datasets, namely LFW, IJB-C and UCCS~\cite{huang2008labeled, maze2018ijbc, guenther2017unconstrained}. 
\rone{We exploit data that do not require domain adaptation to perform gallery specialization, in which the knowledge obtained from networks pre-trained on large face datasets is reused to boost performance on related face recognition tasks.}
We evaluate three architectures for feature extraction, including AFFFE~\cite{li2018eclipse}, a deep-feature extractor adapted to handle misaligned and blurry faces; VGGFace2 SEnet50~\cite{cao2018vggface2}, a backbone that takes advantage of its squeeze-and-excitation blocks; and ArcFace~\cite{deng2019arcface}, a ResNet-101 network that applies a special loss for producing better-suited face representations.

\rtwo{
The proposed approach actually differs from most investigations available in the literature.
To the best of our knowledge, no genuine open-set face recognition work has been evaluated on the IJB-C benchmark. 
Most methods typically aim at improving open-set recognition by providing better feature embeddings for face verification, which comprises a different biometric task.
Moreover, MEL is the first loss function to simultaneously penalize known and negative samples.
Its distinctiveness drives the network toward learning more discriminative face embeddings as it meticulously searches for enhanced parameters.
}

\noindent\textbf{The major contributions of our work are:}
\begin{enumerate}[label=\itshape(\alph*)\upshape]
    \item We evaluate distinct cost functions as well as propose MEL, a novel loss function that maximizes the entropy in order to make training more rigorous. 
    \item We further analyze the Objectosphere loss~\cite{gunther2020watchlist} in favor of verifying how it modifies the feature vector norm of training and test face samples.
    \item We present an adapter network that accelerates the computationally-expensive retraining or fine-tuning of deep convolutional neural networks.
    \item We conduct a detailed open-set analysis of all evaluated cost functions on datasets containing thousands of identities but few samples per class. 
    \item We run experiments to verify whether the proposed approach is effective when combined with distinct deep feature extractors and evaluated on well-known open-set face recognition datasets.
\end{enumerate}

The remainder of this work is organized as follows: 
Section~\ref{sec:related} provides an overview of related work.
Section~\ref{sec:approach} describes the proposed approach: a compact adaptation network combined with MEL or other open-set loss functions.
Section~\ref{sec:experiments} exposes the experimental evaluation on three different face datasets.
Section~\ref{sec:discussion} presents a thorough discussion of the attained results and Section~\ref{sec:conclusion} finishes up with the conclusion and final words.

%% file: sections/related.tex
\section{Related Work}\label{sec:related}
Most modern face recognition systems rely on deep convolutional neural networks (DNNs)~\cite{parkhi2015deep,chen2016unconstrained,liu2017sphereface,wang2018cosface,deng2019arcface,sun2020circle,meng2021magface,kim2022adaface}.
Strategies have been designed to achieve better identification performance on difficult images, such as margin-based or triplet loss, and different network topologies~\cite{schroff2015facenet,sankaranarayanan2016triplet,cao2018vggface2}. 
However, DNNs are not usually designed to handle facial images with a low optical resolution, or even false-positive face detections.
Besides, the aforementioned works do not ``disregard'' low-interest samples and, as a result, end up matching all unknown identities with their respective most similar subjects from the gallery set. 

Vareto~\etal{vareto2017ijcb} combined hashing functions to set up a vote-list histogram. 
Some researchers have adopted \textit{one-vs-all} SVM or PLS models~\cite{chowdhury2016one,dos2014extending} whereas others explored clustering techniques~\cite{henrydoss2020enhancing,vareto2020ijcb}.
The aforementioned methods neither implement the entire closed-set identification pipeline nor comply with the requirements of real-time or real-world applications.
\rtwo{Most present-day methods aim at improving closed-set recognition or person re-identification problems and rarely consider open-set protocols~\cite{liu2017sphereface,wang2018cosface,sun2020circle}.
Others typically focus on open-set recognition by providing better feature embeddings for face verification, which comprises a different biometric task~\cite{deng2019arcface,meng2021magface,kim2022adaface}.}
Moreover, Hassen~\etal{hassen2020learning} introduced a loss function that draws same-class samples near and
Zhou~\etal{zhou2021learning} introduced an additional layer to store class-specific thresholds. 
Researchers have also explored adversarially-generated samples for ``balanced'' decision boundaries among known and unknown classes~\cite{perera2020generative,kong2021opengan,yue2021counterfactual}. 
However, these approaches have been evaluated on datasets holding numerous samples per class and, as a consequence, they are not an accurate portrayal of real-world biometric problems.

Most used datasets in non-face open-set recognition are CIFAR~\cite{krizhevsky2010cifar}, MNIST~\cite{lecun1998cifar}, SVHN~\cite{netzer2011reading} and TinyImageNet, a subset of ImageNet~\cite{deng2009imagenet}, to name a few.
They range from 5 to 20 classes in the known set, but each class encompasses myriads of samples.
Approaches evaluated on such data are not hampered by the shortage of image samples available for training and, in fact, better preserve the inherent data distribution~\cite{cohen1997overfitting}.
Labeled Faces in the Wild (LFW)~\cite{huang2008labeled} used to be the leading facial benchmark.
LFW contains 13,233 images unevenly distributed among almost six thousand classes.
As it was initially designed for verification, experts have proposed non-official open-set protocols~\cite{guenther2017toward,martindez2019shufflefacenet}.
IJB-C~\cite{maze2018ijbc} contains two disjoint gallery partitions of known individuals merged together for closed-set recognition.
The open-set protocol requires the use of a single gallery partition and, hence, half of the probe subjects have no corresponding match in the gallery set.

In contrast to IARPA's benchmarks, the original UnControlled College Students (UCCS) dataset~\cite{sapkota2013largescale} and its extended version~\cite{guenther2017unconstrained} mandate for faces to be detected as part of the recognition pipeline.
The UCCS dataset consists of images captured at a university campus covering different weather conditions.
UCCS's gallery set encompasses 1,085 known subjects, with approximately 20 instances per class, and countless face samples not labeled to any of the known identities.
There are several partially-occluded faces due to lamp posts and tree branches along with accessories like sunglasses, hats, hoodies, or fur jackets that make both detection and recognition in UCCS benchmark a challenging task.

\rtwo{In summary, few works have designed methods to properly tackle open-set face recognition with mechanisms that enable the network to differentiate individuals of interest from unknown people in a scenario with thousands of identities but few samples per class.
With that in mind, we evaluate our proposed approach on realistic face datasets as a meaningful contribution to the biometric discipline.
Due to these fundamental properties and their intrinsic open-set nature, we use both IJB-C and UCCS datasets along with LFW in our experiments.}

%% file: sections/approach.tex
\section{Proposed Approach}
\label{sec:approach}

A watchlist application $S$ generally consists of three sequential stages: $S = S_d \rightarrow S_r \rightarrow S_c$ and should raise an alarm only when probe samples belong to gallery set $G$.
Subsystem $S_d$ corresponds to the face detection and landmark localization method locates faces in the original input image.
For every detected face, the representation module $S_r$ extracts a corresponding numerical feature vector.
The identification subsystem $S_c$ assigns one of the gallery identities $g \in G$ to the probe face sample.
As shown in \fig{pipeline}, we introduce an additional adaptation module $S_a$ that takes original features from the representation stage and further transforms them into attributes that are better suited for the task at hand.

Template $T_g = S_a(S_r(S_d(x_g)))$ corresponds to the mean representation of subject $g$ when multiples sample are available per class.
Similarly, $F_p = S_a(S_r(S_d(x_p)))$ becomes the probe representation.
The classification subsystem $S_c$ computes a similarity score $s(T_g,F_p)$ between $F_p$ and template $T_g$ for each known individual $g \in G$.
Then, $S_c$ rejects probe samples as unknown when they attain scores lower than $\theta$ for every subject of interest.
If not, $F_p$ is assigned to the identity holding highest score $\max_{g \in G}\ s(T_g,F_p)$.

The compact adapter network $S_a$ aims to establish a drastic difference between gallery subjects and unknown faces.
Therefore, it is not possible to enroll new subjects in the gallery set without retraining.
Since we rely on features extracted from representational network $S_r$, retraining the adapter network $S_a$ is fast and can be performed whenever a new subject needs to be enrolled -- given that watchlists are oftentimes relatively stable over time.

\begin{figure}[t!]
  \centering
  \includegraphics[trim={0.01cm 0.01cm 0.01cm 0.01cm}, width=0.93\columnwidth]{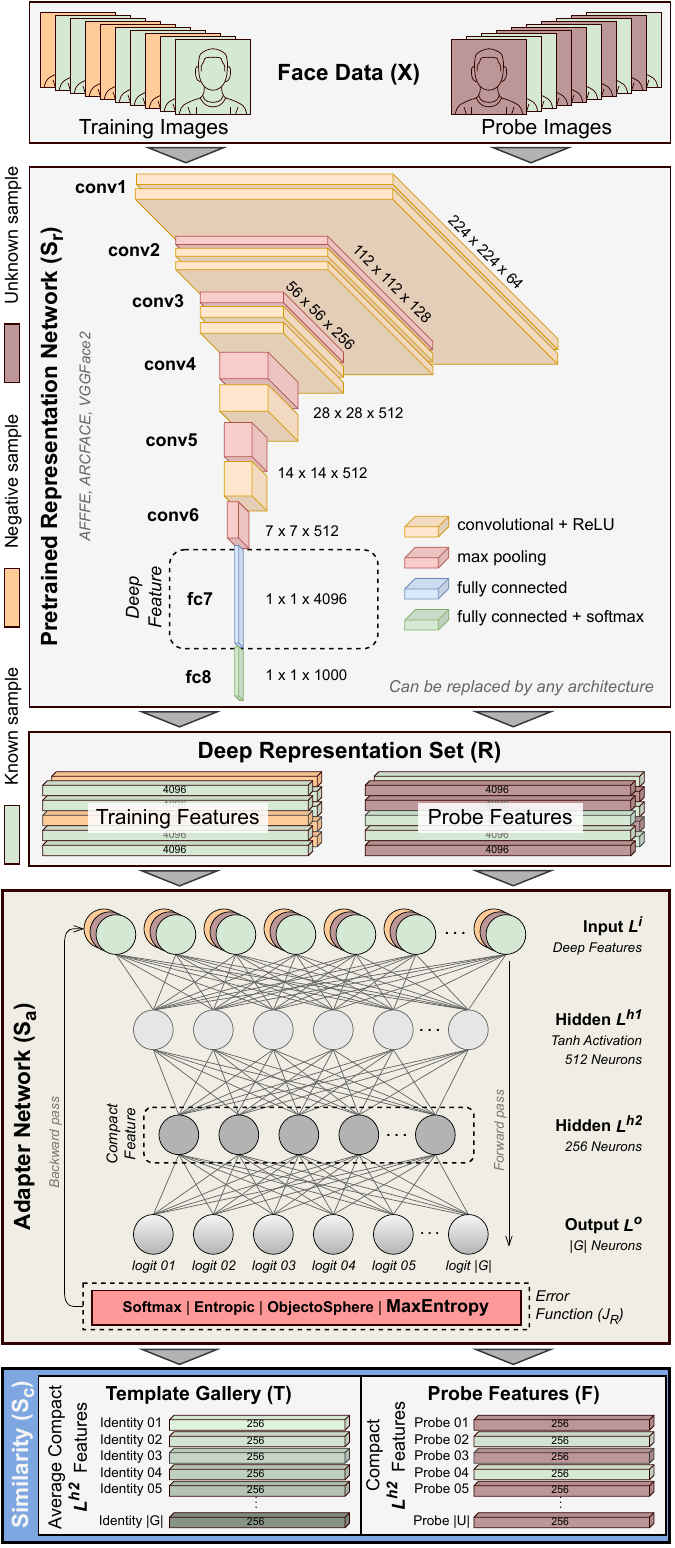}
  \capt{fig:pipeline}{Proposed Pipeline}{
  Given any set of images, training features comprising known and negative samples are input to the proposed adapter network $S_{a}$ for learning parameters that minimize the adopted loss function.
  Compact 256-dimensional features from the penultimate layer of $S_{a}$ are averaged to build a gallery of templates ($T$) during enrollment. On testing, compact features are extracted from probe data and compared with gallery templates $T$ through cosine similarity.}
\end{figure}

\subsection{Training}
One of the fundamental aspects behind the procedure depicted on \fig{pipeline} is that any pre-trained network, such as VGGFace2~\cite{cao2018vggface2}, AFFFE \cite{li2018eclipse} and ArcFace~\cite{deng2019arcface} can be adopted as the pipeline's face representation subsystem $S_r$.
Consequently, the proposed approach does not require time-consuming retraining of massive deep networks every time a new subject is inserted into gallery $G$ since a small adapter network $S_a$ fits the extracted set of representations $R_g=S_r(x_g)~\forall~g \in G$.

\paragraph*{\underline{Adapter Network}}
The adapter network $S_a$ consists of a multi-layer perceptron network with fully-connected layers.
In fact, $S_a$ is composed of an input layer $L^{i}$, two hidden layers $L^{h1}$ and $L^{h2}$, and an output layer $L^{o}$.
The input layer takes in feature vectors $R$ extracted with a pre-trained DNN $S_r$ and, therefore, its size varies according to the deep feature dimension. 
The first hidden layer $L^{h1}$ incorporates a non-linear hyperbolic tangent activation function that outputs values in the range $-1.0$ to $+1.0$ whereas $L^{h2}$ delivers a compact feature representation.

The learning strategy is similar to the training process followed by traditional face recognition systems: we set the output layer $L^{o}$ to hold a size analogous to the number of gallery-enrolled identities.
In other words, each last layer's logit node $L^{o}_{g} \in L^{o}$, also denoted as $l_g \in L^{o}$, stands for the corresponding activation of known subject $g \in G$.
In general, these activations are employed for open-set face classification but they present inferior performance when compared to the distance computation of deep features obtained with neural networks~\cite{gunther2020watchlist}.

The adapter network has been originally designed as a conventional multi-layer perceptron network. 
Ordinarily, its output logit layer $L^{o}$ could be associated with the Softmax activation function $A_{S}$ and assumes a role as the ultimate recognition phase:
\begin{equation}
    \label{eq:softmax}
    A_S(l_g) = \frac{e^{l_g}}{\sum\limits_{1\leq g'\leq |G|}e^{l_{g'}}}
\end{equation} 
However, it performs differently considering that $S_a$ also provides discriminative feature representations that are required in the subsequent similarity classification subsystem $S_c$ (see the blue rectangle in \fig{pipeline}).
The adapter network yields its two last layers during the training stage: 
logits from $L^{h2}$ input Objectosphere while $L^{o}$ values feed the remaining loss functions as detailed below.

\paragraph*{\underline{Entropic Open-Set Loss~($J_E$)}~\cite{dhamija2018objectosphere}}
The Entropic Open-set loss comes to maximize the uncertainty of negative samples by inducing their Softmax responses to lie uniformly distributed.
$J_E$ boosts the maximum entropy distribution of uniform probabilities from negative samples over all $|G|$ known classes registered in the gallery set $G$.
In the classic Cross-Entropy loss, $t_g$ represents a one-hot vector holding the value of one at the index that corresponds to known class $g$.
Under the inclusion of negative instances, $J_E$ attributes uniform values to target vector $\forall g: t_g = \frac{1}{|G|}$ in such a way that unseen samples are considered as equal members of each known identity:
\begin{equation}
    \label{eq:softmax_loss}
    J_E = -\sum_{1\leq g \leq |G|} t_g \log A_S(L^{o}_{g}(R_x))
\end{equation}

\paragraph*{\underline{Maximal Entropy Loss~($J_M$)}}
The proposed Maximal Entropy loss associates the previously stated Entropic Open-set loss with margin-based Softmax ($A_{Sm}$)~\cite{liu2016large,liang2017soft}.
\rthree{Equation~\eqref{eq:soft-margin-softmax} points out how $A_{Sm}$ affixes a non-negative penalty margin $m$ to $A_S$ in order to decrease the intra-class distance and maximize the segregation among distinct classes.
As the penalty increases, a network learns parameters that push samples more firmly toward their class centroids.
The parameter defines a distance among different classes and, consequently, draws same-class samples closer~\cite{liu2016large}.}
\begin{equation}
    A_{Sm}(l_g) = \frac{e^{l_g-m}}{e^{l_g-m} + \sum\limits_{g'\neq g}e^{l_{g'}}}
    \label{eq:soft-margin-softmax}
\end{equation}

The Maximal Entropy Loss $J_M$ combines the best of both worlds since the Soft-Margin Softmax targets known training samples whereas the Entropic Open-set handles negative instances available during the learning stage.
More precisely, function $J_M$ maximizes the entropy regarding the correct target class when $x \in G$ in the interest of making the closed-set identification more rigorous and, as a result, equips the adapter network with more discriminative weights.
The handicap parameter $m$ establishes a decision boundary for a more appropriate separation of known individuals:
\begin{equation}
    J_{M} = 
    \left\{
        \begin{matrix}
            -\log A_{Sm}(L^{o}_{g}(R_x))                                      & \text{if~} x    \in G \\ 
            -\frac{1}{|G|}\sum\limits_{g=1}^{|G|} \log A_S(L^{o}_{g}(R_x))    & \text{if~} x \notin G
        \end{matrix}
    \right.
    \label{eq:maximal-entropy-loss}
\end{equation}

For a negative sample $x \notin G$, the designed loss uniformly distributes the target variable score among all $g \in G$ subjects in an attempt to support the network in distinguishing gallery-enrolled subjects from unknown identities.
Similar to the aforementioned $J_{E}$ loss, the insight of equalizing logit values for unknown samples lies behind not knowing anything about their corresponding identity and, therefore, they hold an equivalent likelihood of being assigned to any subject registered in the gallery set.
Analogous to Dhamija~\etal{dhamija2018objectosphere}, the overall error obtained with $J_{M}$ is minimized when the Softmax responses $A_S(\cdot)$ of negative samples are equally distributed.

\paragraph*{\underline{Objectosphere Loss~($J_O$)}~\cite{dhamija2018objectosphere}}
Objectosphere dissociates representations of known and negative samples by directly modifying their feature magnitudes. 
Since $J_E$ cannot guarantee that such a pattern would be generated for unknown samples, \sfig{teaser:os} illustrates that Objectosphere modifies the network weights to drive negative instances toward the feature space origin.
This is achieved by forcing the magnitude of negative features $||L^{h2}(R_x)||_2$ to be closer to zero while simultaneously pushing known feature magnitudes to at least $\xi$, a required hyperparameter for Objectosphere.
\begin{equation}
    J_O = J_E + \lambda
    \begin{cases}
      \max(\xi - ||L^{h2}(R_x)||_2, 0)^{2}  & \text{if } x \in G\\
      ||L^{h2}(R_x)||_{2}^{2}               & \text{if } x \notin G
    \end{cases}
    \label{eq:objectosphere_loss}
\end{equation}
Larger $\xi$ values scale up deep features, including those extracted from unknown samples, which can be compensated by lower weights in the last layer $L^o$; however, what actually makes a difference is the increased separation among known, negative and, ultimately, unknown samples.

\paragraph*{\underline{Additional Garbage Class~($J_G$)}}
With the high demand for open-set recognition systems and the practicability of the Cross-Entropy loss, accessible in every deep learning framework, a common strategy is to add an extra class $|G|+1$ to encompass negative samples.
We refer to the adapter network $S_a$ trained with $J_G$ as the \texttt{Garbage} approach in the experimentation section.
\begin{equation}
    J_{G} = -\sum\limits_{1\leq g \leq|G|+1} t_g \log A_{S}(L^{o}_{g}(R_x))
    \label{eq:garbage}
\end{equation}

\subsection{Enrollment and Inference}
The enrollment of subjects of interest is illustrated in \fig{pipeline}.
It starts with the extraction of compact features from all gallery samples in the interest of creating a gallery of templates $T$.
Equation~\eqref{eq:template} demonstrates that for each known identity $g \in G$, a unique template $T_g$ is established by averaging the normalized compact features obtained with the adapter network where $|K_g|$ is the number of enrollment samples available for subject $g$.
\begin{equation}
    \label{eq:template}
    T_g = \frac{1}{|K_g|} \sum\limits_{1\leq k \leq K_g}S_a(S_r(S_d(x_{g,k})))
\end{equation}
Analogous feature vectors are obtained for probe images $x_p \in P$ during the inference stage by employing the very same representational and adaptation networks utilized in the enrollment phase:
\begin{equation}
    \label{eq:probe}
    F_p = S_a(S_r(S_d(x_p)))
\end{equation}
Then, the classification module $S_c$ computes similarity scores between probes and all gallery-enrolled identities through the angular cosine similarity:
\begin{equation}
    s(T_g, F_p) = \cos(T_g, F_p)= \frac{{T_g}^{\mathrm{T}} F_p}{||T_g||\cdot||F_p||}
    \label{eq:similarity}
\end{equation}
It is worth mentioning that we have also investigated other similarity-based functions that make use of probe feature magnitudes~\cite{dhamija2018objectosphere,gunther2020watchlist}; however, they include several issues that have not been addressed in this work.

%% file: sections/experiments.tex
\section{Experiments}
\label{sec:experiments}

This section presents the experimental evaluation of the approaches described in Section~\ref{sec:approach}.
It starts detailing the adopted evaluation metrics, assessed methods, and a description of the experimental setup along with the explored datasets.
Further, it provides an experimental assessment of the obtained feature magnitudes and a comparison between the traditional Cross-Entropy and the negative-based cost functions, namely Entropic Open-set, Objectosphere, and the proposed Maximal Entropy Loss.

\subsection{Evaluation Metrics}
We adopt the open-set ROC curve~\cite{phillips2011evaluation,idiap2012beat,grother2022frvt}, which plots the True Positive Identification Rate (TPIR) against the False Positive Identification Rate (FPIR) by varying the rejection threshold $\theta$.
TPIR is computed solely on probe samples of known subjects $\mathbf K$ by considering probes to be correctly identified if the similarity to the correct identity $g^*$ is the highest and above operating threshold $\theta$:
\begin{equation}
    \begin{split}
        \hspace*{-.8em}\mathrm{TPIR}(\theta) = \frac{1}{|\mathbf K|}\Bigl|\bigl\{F_p \in \mathbf K \mid \argmax_{g \in G}\ \cos(T_g,F_p) = g^*\\
        \wedge~ \cos(T_{g^*},F_p) \geq \theta\bigr\}\Bigr|
    \end{split}\hspace*{-.5em}
    \label{eq:dir}
\end{equation}
FPIR corresponds to the false alarm rate triggered by unknown samples $\mathbf U$.
A false positive identification occurs when the similarity of an unknown sample $F_p$ to any of the known subject templates $T_g$ is larger than threshold $\theta$:
\begin{equation}
    \hspace*{-.8em}\mathrm{FPIR}(\theta) = \frac{1}{{\bigl|\mathbf U\bigl|}}\Bigl|\bigl\{F_p \in \mathbf U \mid \max_{g \in G}\ \cos(T_g,F_p) \geq \theta\bigr\}\Bigr|\
    \label{eq:far}
\end{equation}
An optimal open-set face identification system presents a TPIR of $1$ at an FPIR of $0$.
By varying the threshold $\theta$, the open-set ROC curve can be created.

\subsection{Evaluated Datasets}
We utilize a data partition~\cite{guenther2017toward,vareto2020ijcb} that splits LFW into three disjoint groups: 602 known, 1070 negative, and 4096 unknown identities.
We use the provided hand-labeled landmarks available in LFW dataset during the alignment process.
For IJB-C, we train the method on \textit{gallery A} only so that all \textit{gallery B} matching identities available in the probe set act as unknown face samples.
Additionally, LFW is incorporated as the negative set since none of its classes are encountered in IJB-C.
UCCS metadata provides bounding boxes and identity labels, containing either known subject identities or negative labels for unknown faces.
We incorporate the MTCNN face detector~\cite{zhang2016joint} as the default detection system $S_d$ on IJB-C and UCCS benchmarks.
We employ the very same face detector throughout the experiments to standardize the face detection stage.
Following the evaluation protocol, all background detections of MTCNN serve as additional unknown samples during testing in the UCCS dataset \cite{guenther2017unconstrained}.

\subsection{Evaluated Approaches}
In the interest of comparing the proposed adapter network along with Maximal Entropy and Objectosphere loss functions to other methods, we incorporate four additional approaches: \texttt{Baseline}, \texttt{SoftMax}, \texttt{Garbage} and \texttt{Entropic}.
Apart from \texttt{Baseline}, all evaluated methods run the complete pipeline depicted in \fig{pipeline} in which the template gallery consists of feature vectors extracted from the adapter network $S_a$.
\rthree{
In addition to the adapter network, we also investigate whether it is beneficial to fine-tune the entire feature backbone model on the gallery data, which has been shown to be beneficial for larger datasets.
Since this training is much more time consuming, we restrict our experiments to the largest and most difficult dataset, \ie, IJB-C.
}
The \rthree{seven} evaluated techniques are:

\begin{itemize}\itemsep0.0em
    \item \texttt{Baseline} consists of creating a template set with the original features extracted from the representational system $S_r$ and computing the cosine similarity.
    \item\texttt{SoftMax} follows the proposed pipeline by training the adapter network $S_a$ with Cross-Entropy loss, without exploiting any negative samples (negative-free).
    \item \texttt{Garbage} extends \texttt{SoftMax} as it creates a template $T_g$ for each known individual $g \in G$ along with an exclusive template $T_{|G|+1}$ holding negative samples.
    \item \texttt{Entropic} also follows the proposed pipeline, but this time the Entropic Open-set loss is adopted to handle known and negative samples (negative-based).
    \item \texttt{Objectosphere} adopts the Objectosphere loss to train the adapter network with hyperparameters $\xi=1$ and $\lambda=0.01$, as specified in Equation~\eqref{eq:objectosphere_loss}.
    \item \texttt{MaxEntropy} consists of training the adapter network $S_a$ with the proposed Maximal Entropy loss, which holds hyperparameter $m=0.40$ as a default value.
    \item \rthree{\texttt{Finetuning} involves training all layers of the adopted architecture on the evaluated IJB-C dataset.}
\end{itemize}

\begin{figure*}[t!]
    \centering
    \subfloat[\label{fig:lfw:openroc:vgg2}VGGFace2]{\includegraphics[trim={0.1cm 0.0cm 0.0cm 0.5cm},clip,height=4.3cm,page=1]{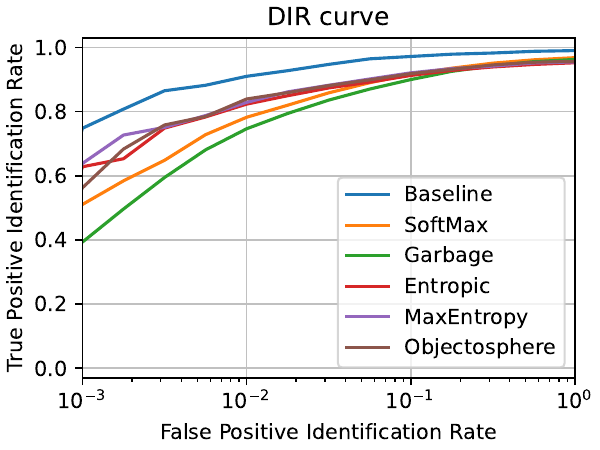}}
    \hspace*{0.01\columnwidth}
    \subfloat[\label{fig:lfw:openroc:afffe}AFFFE]{\includegraphics[trim={0.5cm 0.0cm 0.0cm 0.5cm},clip,height=4.3cm,page=1]{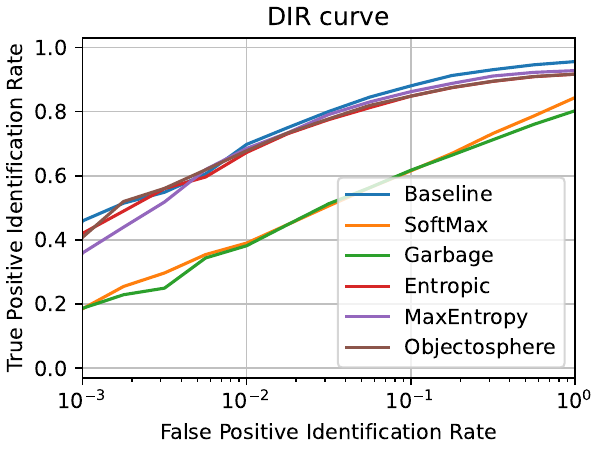}}
    \hspace*{0.01\columnwidth}
    \subfloat[\label{fig:lfw:openroc:arcf}ArcFace]{\includegraphics[trim={0.5cm 0.0cm 0.0cm 0.5cm},clip,height=4.3cm,page=1]{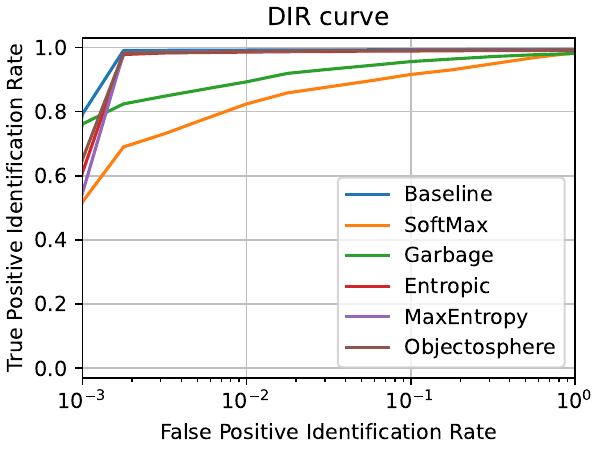}}
    \capt{fig:lfw:openroc}{LFW Evaluation}{Open-set ROC charts are shown for VGGFace2, AFFFE, and ArcFace features.
        Due to the small size of the LFW dataset, FPIR values smaller than $10^{-3}$ cannot reliably be computed and are, hence, left out.
        Because of the small amount of three training samples per identity, the adapter network is not able to provide more meaningful features than the representation network $S_r$.
    }
\end{figure*}

\begin{figure*}[t!]
    \centering
    \subfloat[\label{fig:uccs:openroc:vgg2}VGGFace2]{\includegraphics[trim={0.0cm 0.0cm 0.0cm 0.5cm},clip,height=4.3cm,page=1]{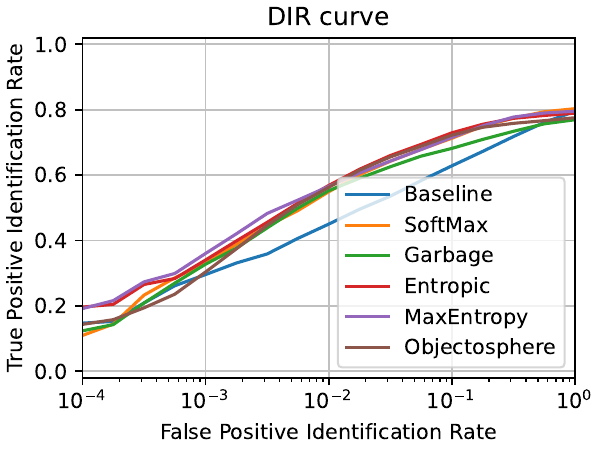}}
    \hspace*{0.01\columnwidth}
    \subfloat[\label{fig:uccs:openroc:afffe}AFFFE]{\includegraphics[trim={0.5cm 0.0cm 0.0cm 0.5cm},clip,height=4.3cm,page=1]{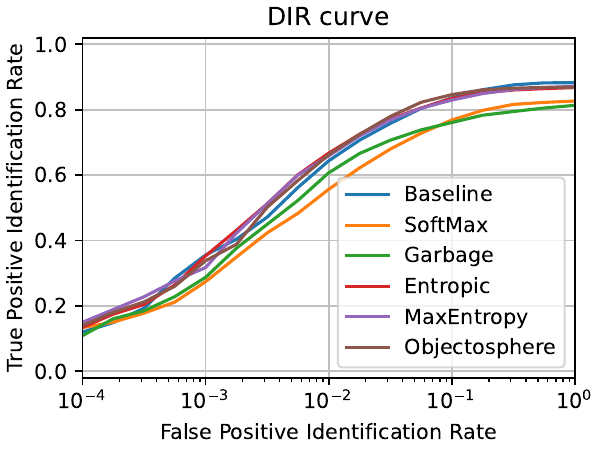}}
    \hspace*{0.01\columnwidth}
    \subfloat[\label{fig:uccs:openroc:arcf}ArcFace]{\includegraphics[trim={0.5cm 0.0cm 0.0cm 0.5cm},clip,height=4.3cm,page=1]{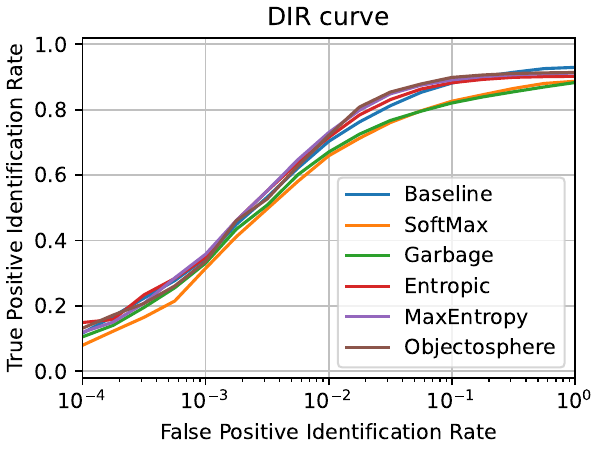}}
    \capt{fig:uccs:openroc}{UCCS evaluation}{
        UCCS is a more challenging dataset than LFW as it comprehends a surveillance and unrestricted domain.
        Therefore, training the adapter network using UCCS known and negative samples improves the performance over the \texttt{Baseline}, especially when training and evaluation samples hold equivalent distribution.
    }
\end{figure*}

\subsection{Network Setup}
The network $S_a$ benefits from representation systems $S_r$, that is, AFFFE, ArcFace and VGGFace2~\cite{li2018eclipse,deng2019arcface,cao2018vggface2} with  1000, 512 and 2048-dimensional deep features, respectively.
The feature extraction counts on Bob's~\cite{anjos2012bob,guenther2012facereclib} biometric pipeline\footnote{\href{https://www.idiap.ch/software/bob/docs/bob/docs/stable/bob/bob.bio.face/doc/baselines.html\#deep-learning-baselines}{https://www.idiap.ch/software/bob/docs/bob/docs/stable/}} that handles face detection, alignment and feature extraction.
The adapter network $S_a$ topology is a compact fully-connected network with $512$ and $256$ neurons in the two hidden layers. 
\rthree{Given the aforestated hyper-parameters, VGGFace2 composes the worst-case scenario in which the adapter network would hold no more than 1.7 million trainable weights, which corresponds to a small fraction of the total of 138 million parameters contained in the deep backbone (98\% less than VGG-16).}

The pipeline is built upon \emph{PyTorch} framework and consists of $500$ training epochs for all datasets. 
Convergence on the validation set was commonly achieved in the first 150 epochs, only minor improvements have been encountered after 200 epochs.
When disregarding the feature extraction process performed in $S_r$, the training procedure takes around 20 minutes for LFW, 80 minutes for UCCS and no more than three hours for IJB-C on a regular multi-core desktop computer with a single \textsc{Nvidia} Titan X GPU.
If more training speed is required, the network topology can be adapted, the number of epochs can be reduced or more GPU resources can be added.

\subsection{Comparison to the State of the Art}\label{sec:comparison}
In the interest of showing the advantage of \texttt{MaxEntropy} and \texttt{Objectosphere} over \texttt{SoftMax}, \texttt{Garbage} and \texttt{Entropic}, the adapter network $S_a$ is trained on different face datasets holding the very same topology and hyper-parameters for all dependent methods.
Figures~\ref{fig:lfw:openroc} through~\ref{fig:ijbc:openroc} depict several approaches in which all of them, except \texttt{Baseline}, rely on $S_a$.
\rtwo{
Additionally, \tab{openroc} provides a detailed list of TPIR values for selected FPIR operating points, evaluated on all three network topologies and all three datasets.
}
The results obtained on the three evaluated datasets are described in the following paragraphs:

\paragraph*{\underline{Labeled Faces in the Wild}}
\fig{lfw:openroc} portrays the investigation on LFW considering different feature representations: VGGFace2, AFFFE and ArcFace.
\texttt{Baseline} presents an outstanding performance using VGGFace2 representation module in \sfig{lfw:openroc:vgg2}, implying that no supplementary data is required for LFW due to its innate characteristics.
Plots \subref*{fig:lfw:openroc:afffe} and \subref*{fig:lfw:openroc:arcf} point out a comparable performance between \texttt{Baseline} and negative-based cost functions.

There is an equivalent behavior with AFFFE when the false-positive proportion exceeds three per thousand samples ($3 \times 10^{-3}$).
ArcFace backbone equipped all approaches with discriminative feature vectors so that very little can be concluded in terms of accuracy. 
Note that four methods attained open-set performance greater than $95\%$ in \subref*{fig:lfw:openroc:arcf} when FPIR surpasses $2 \times 10^{-3}$.
However, results are substantially inferior under \texttt{SoftMax} or \texttt{Garbage} approach.

Unlike most recent face datasets, LFW consists of reasonably good-quality images of cropped faces that cooperate with deep networks in delivering satisfactory feature representations.
As a consequence, computing the cosine distance among original feature vectors, as performed by \texttt{Baseline}, is sufficient to go toward the state of the art.
The small amount of data (three images per subject) seems insufficient to train the adapter network with traditional cost functions.
The adopted non-official protocol~\cite{guenther2017toward} holds nearly 9,300 samples in the probe set and, therefore, the actual threshold value is estimated at no more than 10 images when the FPIR is less than $10^{-3}$.
Moreover, the TPIR performance score for scarce samples is not reliable in low FPIR regions due to the natural threshold fluctuation.

\paragraph*{\underline{UnConstrained College Students}}
\fig{uccs:openroc} discloses the experimental evaluation on the UCCS benchmark.
Along with identities composing the gallery set, UCCS data encompasses both false positive detections (misdetections) and faces from unknown subjects.
\texttt{MaxEntropy} seems capable of attenuating the domain difference between the source data used to train the representation network $S_r$ and the student population scope present in the UCCS dataset.
On the other hand, the domain adaptation seems less impactful for ArcFace features, which indicates that ArcFace architecture can be used in various domains.

\sfig{uccs:openroc:vgg2} reveals that our approach can benefit from the addition of negative samples as the best overall result was achieved with the adapter network when trained with the proposed Maximal Entropy loss.
\sfig{uccs:openroc:afffe} also signalizes significant accuracy gained through the addition of negative samples.
The chart indicates that AFFFE face representations are better adapted for low-resolution images than VGGFace2; however, both are surpassed by ArcFace's robust feature vectors.
\sfig{uccs:openroc:arcf} shows that the negative-exploring cost functions obtain analogous performance: slight dominance of \texttt{MaxEntropy} when FPIR is between $10^{-3}$ and $10^{-1}$.
Although \texttt{Baseline} prevails in the interval $[10^{-1},10^{0}]$, it attains lower accuracy in the aforementioned range along with the other methods.

\begin{figure*}[t!]
    \centering
    \subfloat[\label{fig:ijbc+lfw:openroc:vgg2}VGGFace2]{\includegraphics[trim={0.0cm 0.0cm 0.0cm 0.5cm},clip,height=4.3cm,page=1]{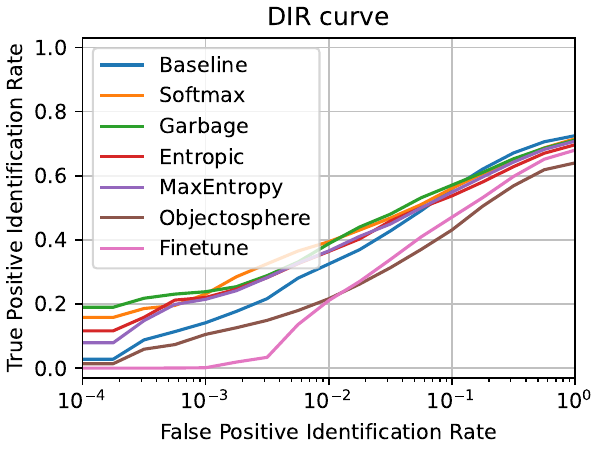}}
    \hspace*{0.01\columnwidth}
    \subfloat[\label{fig:ijbc+lfw:openroc:afffe}AFFFE]{\includegraphics[trim={0.5cm 0.0cm 0.0cm 0.5cm},clip,height=4.3cm,page=1]{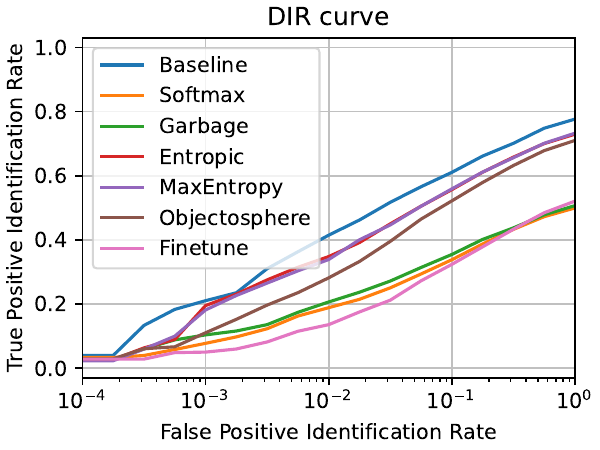}}
    \hspace*{0.01\columnwidth}
    \subfloat[\label{fig:ijbc+lfw:openroc:arcf}ArcFace]{\includegraphics[trim={0.5cm 0.0cm 0.0cm 0.5cm},clip,height=4.3cm,page=1]{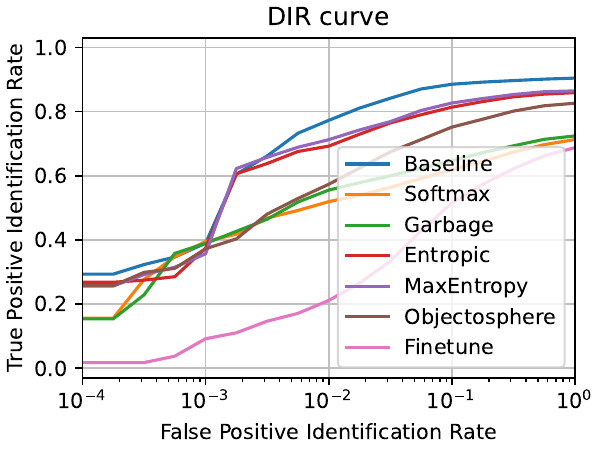}}
    \capt{fig:ijbc+lfw:openroc}{IJB-C + LFW evaluation}{Open-set ROC charts are shown for AFFFE, ArcFace and VGGFace2 features. This evaluation follows IJB-C's open-set protocol test 4 with the addition of the entire LFW dataset as negative samples. Negative data diverging from the gallery set distribution seem incapable to contribute to the method's performance.}
\end{figure*}

\paragraph*{\underline{IARPA Janus Benchmark C series}}
\fig{ijbc+lfw:openroc} exposes experiments on IJB-C merged with more than 13,000 negative samples acquired from LFW.
The discrepancy\footnote{IJB-C contains images without standardized traits whereas LFW comprises mostly good-quality images with close-to-frontal faces.} in image resolution and pose variations between both datasets ends up reflecting on the results as LFW does not play a decisive enhancement role in the proposed adapter network's identification performance when assessing IJB-C benchmark.
\rthree{The three plots suggest that the \texttt{Finetune} approach could not maintain the generalization capability of the original backbone performance either combined with cosine similarity (\texttt{Baseline}) or the adapter network.}

According to \sfig{ijbc+lfw:openroc:vgg2}, negative samples do not seem to provide significant improvement when evaluating VGGFace2 feature vectors and, in fact, they turn out to impair \texttt{Objectosphere}'s exactness.
\sfig{ijbc+lfw:openroc:afffe} corresponds to experiments containing AFFFE features and shows that \texttt{Baseline} outperforms all other approaches.
\texttt{MaxEntropy} attains comparable performance at a low false positive identification rate when it ranges from $1 \times 10^{-3}$ to $3 \times 10^{-3}$.
ArcFace experiments in \sfig{ijbc+lfw:openroc:arcf} also demonstrate the dominance achieved with the \texttt{Baseline} approach.
An approximate accuracy is reached by the \texttt{MaxEntropy} method when FPIR comprises the area to the left of $2 \times 10^{-3}$.

%% file: sections/discussion.tex
\section{Discussion}
\label{sec:discussion}

This section examines the effect of training the adapter network with different-distribution data.
\rtwo{
\tab{openroc} provides a complete view of the results for different FPIRs.
}

\begin{figure*}[t!]
    \vspace{-4.0pt}
    \centering
    \subfloat[\label{fig:ijbc:histograms:or}Representation Network (VGGFace2)]{\includegraphics[trim={1.0cm 0.0cm 1.0cm 1.4cm},clip,width=.31\linewidth,page=1]{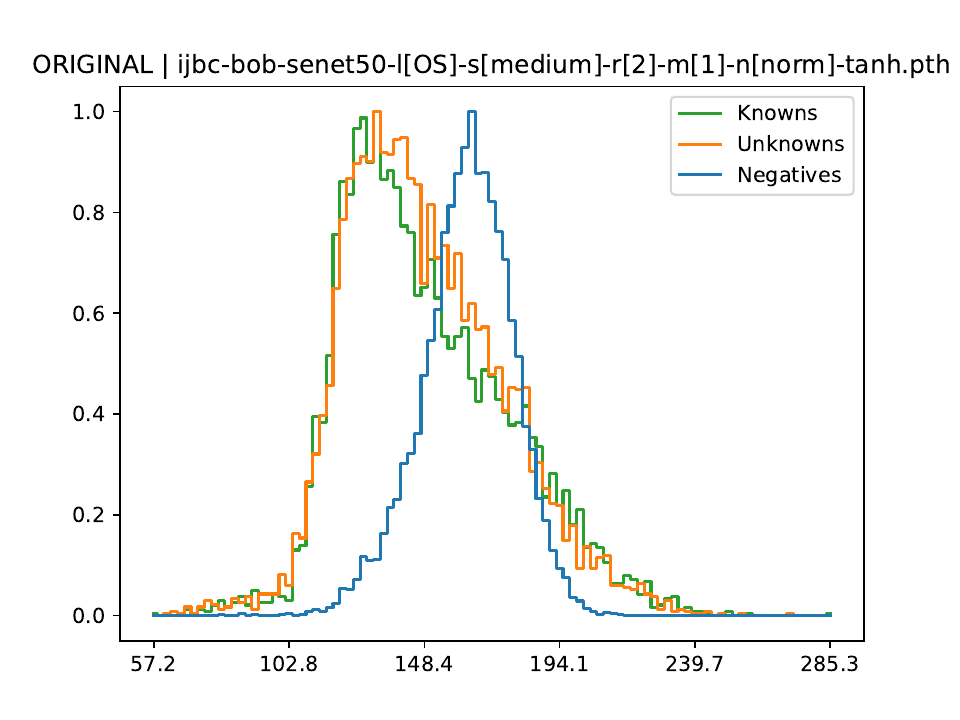}}
    \hspace*{0.02\columnwidth}
    \subfloat[\label{fig:ijbc:histograms:lfw}Negatives from LFW]{\includegraphics[trim={1.0cm 0.0cm 1.0cm 1.4cm},clip,width=.31\linewidth,page=5]{graphics/results/ijbc+lfw/ijbc-bob-senet50-losses.hist}}
    \hspace*{0.02\columnwidth}
    \subfloat[\label{fig:ijbc:histograms:galb}Negatives from IJB-C Galley B]{\includegraphics[trim={1.0cm 0.0cm 1.0cm 1.4cm},clip,width=.31\linewidth,page=9]{graphics/results/ijbc+lfw/ijbc-bob-senet50-losses.hist}}
    \capt{fig:ijbc:histograms}{IJB-C and LFW magnitudes}{
    Chart \subref*{fig:ijbc:histograms:or} exposes training feature magnitudes obtained with VGGFace2 in which knowns and unknowns come from IJB-C and negatives derive from LFW.
    Plots \subref*{fig:ijbc:histograms:lfw} and \subref*{fig:ijbc:histograms:galb} demonstrate how the adapter network $S_a$ combined with Objectosphere behaves on the evaluation data when trained with negatives coming either from LFW or IJB-C gallery B.
    Note that gallery B provides better separation between knowns and unknowns whereas LFW is not sufficient to push the distributions apart.
    }
\end{figure*}

\begin{figure*}[t!]
    \centering
    \subfloat[\label{fig:ijbc:openroc:vgg2}VGGFace2]{\includegraphics[trim={0.0cm 0.0cm 0.0cm 0.5cm},clip,height=4.3cm,page=1]{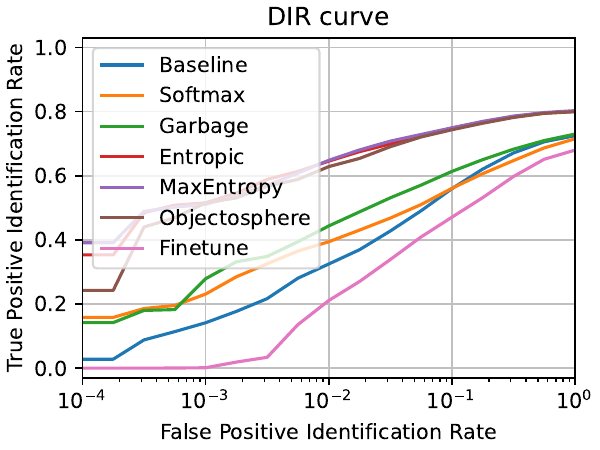}}
    \hspace*{0.01\columnwidth}
    \subfloat[\label{fig:ijbc:openroc:affe}AFFFE]{\includegraphics[trim={0.5cm 0.0cm 0.0cm 0.5cm},clip,height=4.3cm,page=1]{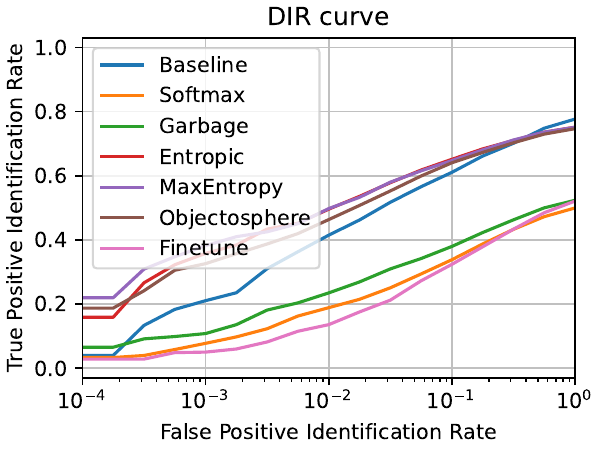}}
    \hspace*{0.01\columnwidth}
    \subfloat[\label{fig:ijbc:openroc:arcf}ArcFace]{\includegraphics[trim={0.5cm 0.0cm 0.0cm 0.5cm},clip,height=4.3cm,page=1]{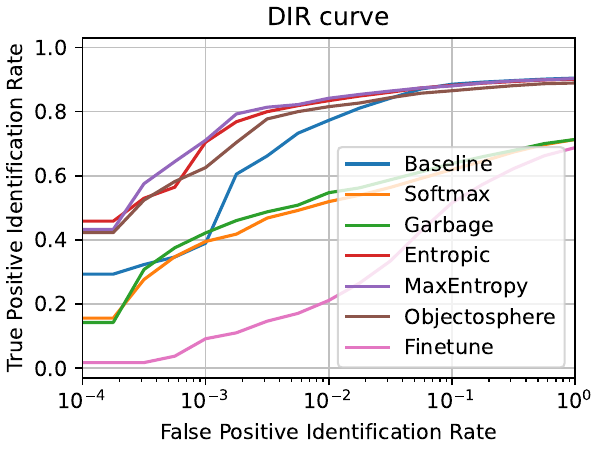}}
    \capt{fig:ijbc:openroc}{IJB-C evaluation}{Open-set ROC charts are shown for AFFFE, ArcFace and VGGFace2 features. Negative samples are obtained from gallery set B of IJB-C dataset. This evaluation does not adhere to IJB-C's open-set protocol test 4.
    When negative samples embody the same distribution as known samples, ``negative-based'' cost functions along with the adapter network outperform the \texttt{Baseline}.
    }
    \vspace{0.5cm}
\end{figure*}

\subsection{Differences between IJB-C and LFW}
Dhamija~\etal{dhamija2018objectosphere} pointed out that the choice of negative samples plays an important role when training an open-set network.
\sfig{ijbc:histograms:or} shows that LFW does not follow the same feature distribution as IJB-C.
As revealed in \fig{ijbc+lfw:openroc}, selecting LFW to compose the set of negative samples could not provide further improvements and outperform the baseline method on IJB-C dataset except for experiments containing VGGFace2 representations.

IJB-C probe samples as well as its enrollment data are distributed differently.
More precisely, gallery-enrolled samples contain mostly good-quality still photos whereas probe samples are mainly composed of low-resolution still images or blurred video frames.
We tend to believe that the adapter network $S_a$ over-adapts to good-quality enrollment samples when it inputs only high-standard data.
Therefore, module $S_a$ ends up lowering the performance on IJB-C by rejecting many probe samples as unknown.
Selecting enrollment and probe data with similar distribution is likely to increase performance.

\fig{ijbc:openroc} discloses an additional set of experiments on IJB-C benchmark in which \textit{gallery A} populates the known set and \textit{gallery B} composes the negative set for \texttt{Entropic}, \texttt{MaxEntropy} and \texttt{Objectosphere}.
This scenario affords a related data distribution between both training subsets.
Despite probe and enrollment data sharing different capture quality, results show that appropriate negative samples significantly improve the open-set face recognition pipeline.
All charts indicate a dominance of \texttt{MaxEntropy} over negative-free methods when FPIR lies below $10^{-1}$.
Factually, using ArcFace backbone achieves the highest accuracy rate of all experiments conducted on IJB-C dataset.

We reckon that real-world watchlist applications would scarcely ever contain negative identities overlapping with unknown face samples.
However, the assessment displayed in \fig{ijbc:openroc} provides a reference point on the maximal identification correctness.
Results show a recurring superiority of \texttt{MaxEntropy} regardless of the adopted representation network.
Unlike \fig{ijbc+lfw:openroc}, where gallery and negative samples hold contrasting data distribution, the resemblance between both IJB-C disjoint galleries delivers discriminative class boundaries.
Distribution-alike data is a must-have aspect required by ``negative-based'' error functions when seeking negative samples in exchange for a meaningful contribution to the open-set recognition pipeline.

Although experiments showed in \fig{ijbc:openroc} do not adhere to the official IJB-C protocol, there are scenarios in which this training scheme would be appropriate.
For instance, an enterprise may have premium clients that must be treated differently than regular customers.
They could be addressed by name and offered a comfortable room on the premises.
Privileged customers constitute known classes but the remaining ones are placed in the negative set.
Prospect customers (unknowns) lie somewhere in between and shall be treated better than the ordinary, but not as good as premium.
Consequently, the face recognition system is supposed to raise an alert whenever premium customers come over.

\begin{table*}[!t]
\centering
\footnotesize
\capt{tab:openroc}{Open-set ROC evaluation}{\rtwo{Open-set ROC results are shown for AFFFE, ArcFace and VGGFace2 feature representations on the three evaluated datasets, namely LFW, UCCS and IJB-C. Each cell consists of the True-Positive Identification Rate where False-Positive Identification Rate value is indicated in the first column (TPIR@FPIR). Best values per model are highlighted in bold, second in italics.}}
\begin{tabular}{c|r||lll|lll|lll}\hline
\textbf{O-ROC }            & \mcbf{1}{r}{Datasets:} & \mcbf{3}{c}{LFW}              & \mcbf{3}{c}{UCCS}             & \mcbf{3}{c}{IJB-C}          \\\hline
FPIR                       & Methods                & VGG2    & AFFFE   & ArcFace   & VGG2    & AFFFE   & ArcFace   & VGG2    & AFFFE   & ArcFace \\\hline\hline
\multirow{6}{*}{$10^{0}$}  & Baseline               & \bf0.99128 & \bf0.99128 & \bf0.99631   & 0.79398 & \bf0.88193 & \bf0.92892   & \bf0.72426 & \bf0.77669 & \bf0.90307 \\
                           & SoftMax                & \it0.96887 & \it0.96887 & 0.98649   & \bf0.80361 & 0.82651 & 0.88795   & 0.68137 & 0.47853 & 0.66994 \\
                           & Garbage                & 0.96264 & 0.96264 & 0.98034   & 0.76747 & 0.81325 & 0.88313   & 0.67525 & 0.40368 & 0.61595 \\
                           & Entropic               & 0.95268 & 0.95268 & \bf0.99631   & 0.79036 & 0.86506 & 0.90000   & 0.69240 & \it0.73374 & 0.85521 \\
                           & Objectosphere          & 0.95143 & 0.95143 & \bf0.99631   & 0.77831 & \it0.87108 & 0.91205   & 0.67034 & 0.70920 & 0.82699 \\
                           & MaxEntropy             & 0.95641 & 0.95641 & \bf0.99631   & \it0.79639 & \it0.87108 & \it0.91566   & \it0.70711 & 0.73006 & \it0.86626 \\\hline

\multirow{6}{*}{$10^{-1}$} & Baseline               & \bf0.97136 & \bf0.87975 & \bf0.99386   & 0.62651 & 0.83253 & 0.88313   & \bf0.56127 & \bf0.61104 & \bf0.88466 \\
                           & SoftMax                & 0.91158 & 0.61595 & 0.91646   & 0.71084 & 0.76988 & 0.82771   & 0.52574 & 0.31043 & 0.57301 \\
                           & Garbage                & 0.89788 & 0.61595 & 0.95577   & 0.68193 & 0.76145 & 0.82048   & 0.52083 & 0.23926 & 0.49939 \\
                           & Entropic               & 0.91283 & 0.84540 & 0.99017   & \bf0.73253 & 0.83012 & 0.87952   & 0.53064 & 0.55092 & 0.81227 \\
                           & Objectosphere          & \it0.92030 & 0.84540 & 0.99140   & 0.71807 & \bf0.84819 & \bf0.90000   & 0.47181 & 0.51656 & 0.73865 \\
                           & MaxEntropy             & \it0.92030 & \it0.86258 & \bf0.99386   & \it0.71928 & \it0.83976 & \it0.89398   & \it0.54289 & \it0.55951 & \it0.82454 \\\hline

\multirow{6}{*}{$10^{-2}$} & Baseline               & \bf0.90909 & \bf0.69693 & \bf0.99509   & 0.44940 & 0.64337 & 0.70120   & 0.32230 & \bf0.41227 & \bf0.77178 \\
                           & SoftMax                & 0.78207 & 0.39018 & 0.82310   & 0.54699 & 0.55783 & 0.66024   & 0.32598 & 0.17546 & 0.44908 \\
                           & Garbage                & 0.74595 & 0.38160 & 0.89312   & 0.55060 & 0.60482 & 0.67108   & 0.33211 & 0.13129 & 0.38773 \\
                           & Entropic               & 0.82067 & 0.66748 & 0.98526   & \bf0.57229 & \bf0.66988 & 0.71325   & \bf0.37745 & 0.33742 & 0.69202 \\
                           & Objectosphere          & \it0.83935 & 0.67853 & 0.98894   & 0.56386 & 0.66024 & \it0.72410   & 0.26348 & 0.27853 & 0.53988 \\
                           & MaxEntropy             & 0.82939 & \it0.68466 & \it0.99017   & \it0.56747 & \it0.66627 & \bf0.73253   & \it0.37500 & \it0.34479 & \it0.70552 \\\hline

\multirow{6}{*}{$10^{-3}$} & Baseline               & \bf0.74720 & \bf0.45767 & \bf0.79238   & 0.29277 & \bf0.35663 & \it0.35060   & 0.14093 & \bf0.21104 & \it0.39387 \\
                           & SoftMax                & 0.51059 & 0.18405 & 0.51966   & 0.33614 & 0.26988 & 0.31325   & \it0.22304 & 0.09816 & 0.34969 \\
                           & Garbage                & 0.38979 & 0.18405 & \it0.76290   & 0.32651 & 0.28916 & 0.32651   & 0.20221 & 0.05276 & 0.28221 \\
                           & Entropic               & 0.62889 & \it0.42209 & 0.61794   & \it0.34337 & \it0.35301 & 0.34337   & 0.21814 & 0.17301 & 0.36319 \\
                           & Objectosphere          & 0.56663 & 0.41104 & 0.66216   & 0.30120 & 0.33976 & 0.33614   & 0.16299 & 0.08834 & 0.38405 \\
                           & MaxEntropy             & \it0.64010 & 0.35583 & 0.55283   & \bf0.35904 & 0.31687 & \bf0.36145   & \bf0.23897 & \it0.18650 & \bf0.40982 \\\hline
\end{tabular}
\end{table*}

\subsection{Deep Feature Magnitudes}
Objectosphere loss aims to push feature magnitude extracted from unknown samples to very low figures.
It simultaneously attempts to shift the magnitude of known samples toward a specified value $\xi$.
Robust open-set methods are expected to achieve high accuracy in different datasets with consistent parameters.
This requirement plays an essential role in biometrics as it is not possible to anticipate the visual traits of all probe samples.
Best results have been attained on UCCS when setting \texttt{Objectosphere} parameters $\xi=1$ and $\lambda=0.01$, and we have verified that these parameters also work well on LFW and IJB-C.

\begin{figure*}[t!]
    \vspace{-5.0pt}
    \centering
    \subfloat[\label{fig:uccs:histograms:or}Representation Network (VGGFace2)]{\includegraphics[trim={1.0cm 0.0cm 1.0cm 1.4cm},clip,width=.30\linewidth,page=1]{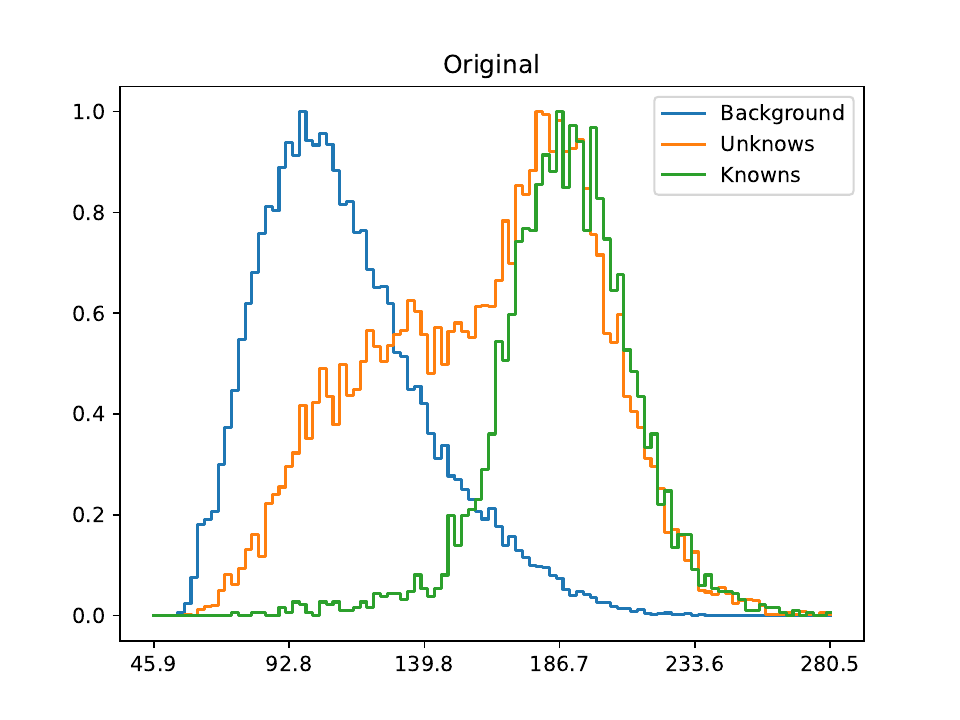}}
    \hspace*{0.02\columnwidth}
    \subfloat[\label{fig:uccs:histograms:ce}Adapter Network + SoftMax]{\includegraphics[trim={1.0cm 0.0cm 1.0cm 1.4cm},clip,width=.30\linewidth,page=2]{graphics/results/uccs/vgg2-long}}
    \hspace*{0.02\columnwidth}
    \subfloat[\label{fig:uccs:histograms:os}Adapter Network + Objectosphere]{\includegraphics[trim={1.0cm 0.0cm 1.0cm 1.4cm},clip,width=.30\linewidth,page=5]{graphics/results/uccs/vgg2-long}}
    \capt{fig:uccs:histograms}{UCCS magnitudes}{
    All plots portray results obtained with probe data:
    \textbf{Knowns} designates subjects registered in the watchlist, \textbf{Unknowns} specifies probe samples without corresponding identity in the gallery set, and \textbf{Background} refers to face misdetections.
    Feature magnitude is an indicator that domain adaptation improves separation between known and unknown subjects.
    While SoftMax provides low magnitudes for background and approximately half of the unknown samples, Objectosphere delivers even better separation.
    }
\end{figure*}

\fig{uccs:histograms} displays deep feature magnitude histograms for UCCS evaluation data.
Original VGGFace2 features hold a considerable magnitude overlap among unknown and known subjects as well as false-positive detections in the background.
The intersection remains when training the adapter network with \texttt{SoftMax}, but \texttt{Objectosphere} reduces the coincidental area between known and unknown samples.
Basically, weights learned with \texttt{Objectosphere} can distinguish enrolled subjects from unknown identities during the testing stage.
Known samples are distributed closer to the desired target magnitude whereas negative (background) samples have a peak close to 0, but are distributed throughout the range of magnitudes.

As indicated in \sfig{ijbc:histograms:or}, LFW images provide higher magnitudes but IJB-C instances result in low-magnitude representations.
Deep networks may misclassify probe samples since the image quality has an impact on the acquired feature vectors.
Due to the lack of similarity between IJB-C and LFW, the latter is not capable to guide the adapter network $S_a$ in discriminating IJB-C probe samples.
\sfig{ijbc:histograms:lfw} and \subref*{fig:ijbc:histograms:galb} present probe feature magnitudes when $S_a$ is trained with negatives proceeding from LFW or IJB-C's \textit{gallery B}.
Note that the magnitudes are well-above the intended separation threshold $\xi=1$ and, hence, appropriate negatives might help to separate further.

\subsection{Proposed Approach Applicability}
\texttt{MaxEntropy} requires a distance margin $m$ in the interval $[0,1]$ whereas \texttt{Objectosphere} includes sphere-related and regularizing parameters ($\xi$ and $\lambda$, respectively).
As a result, associating both losses culminates in the specification of three hyper-parameters, not including the ones regarding the adaptation network, such as the number of neurons, learning rate, and batch size to name a few.
Cost functions that require the adjustment of multiple parameters make their deployment unfeasible in both academic and realistic scenarios.
Consequently, we do not combine \texttt{MaxEntropy} with \texttt{Objectosphere} since a large number of tunable parameters turns into an optimization problem.

We acknowledge that a desirable open-set face recognition approach would only require the enrollment of subjects of interest without the need to fine-tune the deep representation backbone.
However, the three evaluated datasets are composed of numerous identities holding very few samples per class in the training set, a common requisite in watchlist problems.
Since applying an untouched pre-trained representation model to dissimilar data distributions regularly results in a substantial accuracy loss, the designed adapter network $S_a$ offers a flexible trade-off between computational time and correctness.

%% file: sections/conclusion.tex
\section{Conclusion}\label{sec:conclusion}

Pre-trained deep networks usually require considerable time to be adapted and retrained to new domains, especially when the training data is constantly updated.
This is the scenario in which the proposed compact adapter network comes in handy as it serves as a quick-trainable replacement for the output layer.
Moreover, the evaluated cost functions take advantage of supplementary information when adding negative samples to the training stage.
Experiments have shown that additional samples play an important role in ``identifying'' the unknown when the training samples are sufficiently representative of the uninvestigated feature space.

The proposed approach is suited for watchlists and transfer-learning tasks since the adaptation network can be attached to the output of any pre-trained deep network model and be quickly adjusted to different data distributions.
Retraining large deep backbones, such as ArcFace and VGGFace2, every time a new identity is added to the gallery set becomes categorically infeasible \rthree{and has proven to be counterproductive}.
The ArcFace network, for instance, contains nearly 50 million weights in contrast to 394,850 parameters in the adapter network when trained on LFW and inputting 512-dimensional feature embeddings from ArcFace/ResNet-100.

Experiments carried out on the open-set face recognition protocols of LFW, UCCS, and IJB-C have provided a comprehensive analysis of the compact network and the employed loss functions.
The evaluation has shown that the association of the adapter network with Objectosphere or the proposed Maximal Entropy loss is capable of outperforming the original deep features in many cases.
As detailed in the literature comparison, part of the adopted negative images clearly boosted the performance of our method whereas others encompassing distinct domains as well as different data distribution were not adequate and contributed little to the approach accuracy.
How to obtain or generate more effective negative samples will be investigated in future work.

\section*{Acknowledgments} 
We are thankful to the Brazilian National Council for Scientific and Technological Development -- CNPq (Grants 309953/2019-7 and 203402/2020-0), the Minas Gerais Research Foundation -- FAPEMIG (Grant~PPM-00540-17), the Federal University of Minas Gerais (UFMG) and, especially, the University of Z\"urich (UZH).


%% file: main_IVC.bbl
\begin{thebibliography}{10}

\bibitem{hill2020wrongfully}
Kashmir Hill.
\newblock Wrongfully accused by an algorithm.
\newblock {\em New York Times}, 6 2020.
\newblock
  \url{https://www.nytimes.com/2020/06/24/technology/facial-recognition-arrest.html}.

\bibitem{gunther2020watchlist}
Manuel G\"unther, Akshay~Raj Dhamija, and Terrance~E. Boult.
\newblock Watchlist adaptation: Protecting the innocent.
\newblock In {\em International Conference of the Biometrics Special Interest
  Group (BIOSIG)}, 2020.

\bibitem{romm2017amazons}
Tony Romm.
\newblock Amazon’s facial-recognition tool misidentified 28 lawmakers as
  people arrested for a crime, study finds.
\newblock {\em Washington Post}, July 2017.

\bibitem{wen2016centerloss}
Yandong Wen, Kaipeng Zhang, Zhifeng Li, and Yu~Qiao.
\newblock A discriminative feature learning approach for deep face recognition.
\newblock In {\em European Conference on Computer Vision (ECCV)}. Springer,
  2016.

\bibitem{dhamija2018objectosphere}
Akshay~Raj Dhamija, Manuel G\"unther, and Terrance~E. Boult.
\newblock Reducing network agnostophobia.
\newblock In {\em Advances in Neural Information Processing Systems (NeurIPS},
  2018.

\bibitem{palechor2023protocols}
Andres Palechor, Annesha Bhoumik, and Manuel G\"unther.
\newblock Large-scale open-set classification protocols for imagenet.
\newblock In {\em Winter Conference on Applications of Computer Vision (WACV)}.
  CVF/IEEE, January 2023.

\bibitem{huang2008labeled}
Gary~B. Huang, Manu Ramesh, Tamara Berg, and Erik Learned-Miller.
\newblock Labeled faces in the wild: A database for studying face recognition
  in unconstrained environments.
\newblock Technical Report 07-49, Univ. of Massachusetts, Amherst, 2007.

\bibitem{maze2018ijbc}
Brianna Maze, Jocelyn Adams, James~A. Duncan, Nathan Kalka, Tim Miller, Charles
  Otto, Anil~K. Jain, W.~Tyler Niggel, Janet Anderson, Jordan Cheney, and
  Patrick Grother.
\newblock {IARPA} {Janus} {Benchmark - C}: Face dataset and protocol.
\newblock In {\em International Conference on Biometrics (ICB)}, 2018.

\bibitem{guenther2017unconstrained}
Manuel G\"unther, Peiyun Hu, Christian Herrmann, Chi-Ho Chan, Min Jiang, Shufan
  Yang, Akshay~Raj Dhamija, Deva Ramanan, J{\"u}rgen Beyerer, Josef Kittler,
  et~al.
\newblock Unconstrained face detection and open-set face recognition challenge.
\newblock In {\em International Joint Conference on Biometrics (IJCB)}. IEEE,
  2017.

\bibitem{li2018eclipse}
Chunchun Li, Manuel G\"unther, and Terrance~E. Boult.
\newblock {ECLIPSE}: Ensembles of centroids leveraging iteratively processed
  spatial eclipse clustering.
\newblock In {\em Winter Conference on Applications of Computer Vision (WACV)},
  2018.

\bibitem{cao2018vggface2}
Qiong Cao, Li~Shen, Weidi Xie, Omkar~M. Parkhi, and Andrew Zisserman.
\newblock {VGGFace2}: A dataset for recognising faces across pose and age.
\newblock In {\em Automatic Face {\&} Gesture Recognition (FG)}. IEEE, 2018.

\bibitem{deng2019arcface}
Jiankang Deng, Jia Guo, Niannan Xue, and Stefanos Zafeiriou.
\newblock {ArcFace}: Additive angular margin loss for deep face recognition.
\newblock In {\em Conference on Computer Vision and Pattern Recognition
  (CVPR)}, 2019.

\bibitem{parkhi2015deep}
Omkar~M Parkhi, Andrea Vedaldi, and Andrew Zisserman.
\newblock Deep face recognition.
\newblock In {\em British Machine Vision Conference (BMVC)}, 2015.

\bibitem{chen2016unconstrained}
Jun-Cheng Chen, Vishal~M Patel, and Rama Chellappa.
\newblock Unconstrained face verification using deep {CNN} features.
\newblock In {\em Winter Conference on Applications of Computer Vision (WACV)},
  2016.

\bibitem{liu2017sphereface}
Weiyang Liu, Yandong Wen, Zhiding Yu, Ming Li, Bhiksha Raj, and Le~Song.
\newblock Sphereface: Deep hypersphere embedding for face recognition.
\newblock In {\em Proceedings of the IEEE/CVF conference on computer vision and
  pattern recognition}, 2017.

\bibitem{wang2018cosface}
Hao Wang, Yitong Wang, Zheng Zhou, Xing Ji, Dihong Gong, Jingchao Zhou, Zhifeng
  Li, and Wei Liu.
\newblock Cosface: Large margin cosine loss for deep face recognition.
\newblock In {\em Proceedings of the IEEE/CVF conference on computer vision and
  pattern recognition}, 2018.

\bibitem{sun2020circle}
Yifan Sun, Changmao Cheng, Yuhan Zhang, Chi Zhang, Liang Zheng, Zhongdao Wang,
  and Yichen Wei.
\newblock Circle loss: A unified perspective of pair similarity optimization.
\newblock In {\em Proceedings of the IEEE/CVF conference on computer vision and
  pattern recognition}, 2020.

\bibitem{meng2021magface}
Qiang Meng, Shichao Zhao, Zhida Huang, and Feng Zhou.
\newblock {MagFace}: A universal representation for face recognition and
  quality assessment.
\newblock In {\em Conference on Computer Vision and Pattern Recognition
  (CVPR)}, 2021.

\bibitem{kim2022adaface}
Minchul Kim, Anil~K Jain, and Xiaoming Liu.
\newblock Adaface: Quality adaptive margin for face recognition.
\newblock In {\em Conference on Computer Vision and Pattern Recognition
  (CVPR)}, 2022.

\bibitem{schroff2015facenet}
Florian Schroff, Dmitry Kalenichenko, and James Philbin.
\newblock Facenet: A unified embedding for face recognition and clustering.
\newblock In {\em Conference on Computer Vision and Pattern Recognition
  (CVPR)}, 2015.

\bibitem{sankaranarayanan2016triplet}
Swami Sankaranarayanan, Azadeh Alavi, Carlos~D. Castillo, and Rama Chellappa.
\newblock Triplet probabilistic embedding for face verification and clustering.
\newblock In {\em Biometrics Theory, Applications and Systems (BTAS)}. IEEE,
  2016.

\bibitem{vareto2017ijcb}
Rafael~Henrique Vareto, Samira Silva, Filipe Costa, and William~Robson
  Schwartz.
\newblock Towards open-set face recognition using hashing functions.
\newblock In {\em International Joint Conference on Biometrics (IJCB)}, 2017.

\bibitem{chowdhury2016one}
Aruni~Roy Chowdhury, Tsung-Yu Lin, Subhransu Maji, and Erik Learned-Miller.
\newblock One-to-many face recognition with bilinear {CNNs}.
\newblock In {\em Winter Conference on Applications of Computer Vision (WACV)},
  2016.

\bibitem{dos2014extending}
Cassio~Elias Dos~Santos and William~Robson Schwartz.
\newblock Extending face identification to open-set face recognition.
\newblock In {\em Brazilian Symposium on Computer Graphics and Image Processing
  (SIBGRAPI)}, 2014.

\bibitem{henrydoss2020enhancing}
James Henrydoss, Steve Cruz, Chunchun Li, Manuel G\"unther, and Terrance~E.
  Boult.
\newblock Enhancing open-set recognition using clustering-based extreme value
  machine ({C-EVM}).
\newblock In {\em International Conference on Big Data (BigData)}. IEEE, 2020.

\bibitem{vareto2020ijcb}
Rafael~Henrique Vareto and William~Robson Schwartz.
\newblock Unconstrained face identification using ensembles trained on
  clustered data.
\newblock In {\em International Joint Conference on Biometrics (IJCB)}, 2020.

\bibitem{hassen2020learning}
Mehadi Hassen and Philip~K Chan.
\newblock Learning a neural-network-based representation for open set
  recognition.
\newblock In {\em International Conference on Data Mining}. SIAM, 2020.

\bibitem{zhou2021learning}
Da-Wei Zhou, Han-Jia Ye, and De-Chuan Zhan.
\newblock Learning placeholders for open-set recognition.
\newblock In {\em Conference on Computer Vision and Pattern Recognition
  (CVPR)}, 2021.

\bibitem{perera2020generative}
Pramuditha Perera, Vlad~I Morariu, Rajiv Jain, Varun Manjunatha, Curtis
  Wigington, Vicente Ordonez, and Vishal~M. Patel.
\newblock Generative-discriminative feature representations for open-set
  recognition.
\newblock In {\em Conference on Computer Vision and Pattern Recognition
  (CVPR)}, 2020.

\bibitem{kong2021opengan}
Shu Kong and Deva Ramanan.
\newblock Opengan: Open-set recognition via open data generation.
\newblock In {\em International Conference on Computer Vision (ICCV)}, 2021.

\bibitem{yue2021counterfactual}
Zhongqi Yue, Tan Wang, Qianru Sun, Xian-Sheng Hua, and Hanwang Zhang.
\newblock Counterfactual zero-shot and open-set visual recognition.
\newblock In {\em Conference on Computer Vision and Pattern Recognition
  (CVPR)}, 2021.

\bibitem{krizhevsky2010cifar}
Alex Krizhevsky, Vinod Nair, and Geoffrey Hinton.
\newblock Cifar datasets-10.
\newblock URL http://www.cs.toronto.edu/kriz/cifar.html, 2010.

\bibitem{lecun1998cifar}
Yann LeCun, Corinna Cortes, and Christopher~J.C. Burges.
\newblock Mnist handwritten digits dataset.
\newblock URL http://yann.lecun.com/exdb/mnist/index.html, 1998.

\bibitem{netzer2011reading}
Yuval Netzer, Tao Wang, Adam Coates, Alessandro Bissacco, Bo~Wu, and Andrew~Y
  Ng.
\newblock Reading digits in natural images with unsupervised feature learning.
\newblock In {\em Advances in Neural Information Processing Systems (NeurIPS)
  Workshop}, 2011.

\bibitem{deng2009imagenet}
Jia Deng, Wei Dong, Richard Socher, Li-Jia Li, Kai Li, and Li~Fei-Fei.
\newblock Imagenet: A large-scale hierarchical image database.
\newblock In {\em Conference on Computer Vision and Pattern Recognition
  (CVPR)}. IEEE, 2009.

\bibitem{cohen1997overfitting}
Paul~R Cohen and David Jensen.
\newblock Overfitting explained.
\newblock In {\em International Workshop on Artificial Intelligence and
  Statistics}. PMLR, 1997.

\bibitem{guenther2017toward}
Manuel G{\"u}nther, Steve Cruz, Ethan~M. Rudd, and Terrance~E. Boult.
\newblock Toward open-set face recognition.
\newblock In {\em Conference on Computer Vision and Pattern Recognition (CVPR)
  Workshops}, 2017.

\bibitem{martindez2019shufflefacenet}
Yoanna Martindez-Diaz, Luis~S Luevano, Heydi Mendez-Vazquez, Miguel
  Nicolas-Diaz, Leonardo Chang, and Miguel Gonzalez-Mendoza.
\newblock Shufflefacenet: A lightweight face architecture for efficient and
  highly-accurate face recognition.
\newblock In {\em International Conference on Computer Vision (ICCV)
  Workshops}, 2019.

\bibitem{sapkota2013largescale}
Archana Sapkota and Terrance~E. Boult.
\newblock Large scale unconstrained open set face database.
\newblock In {\em Biometrics Theory, Applications and Systems (BTAS)}, 2013.

\bibitem{liu2016large}
Weiyang Liu, Yandong Wen, Zhiding Yu, and Meng Yang.
\newblock Large-margin softmax loss for convolutional neural networks.
\newblock In {\em International Conference on Machine Learning}, pages
  507--516. PMLR, 2016.

\bibitem{liang2017soft}
Xuezhi Liang, Xiaobo Wang, Zhen Lei, Shengcai Liao, and Stan~Z Li.
\newblock Soft-margin softmax for deep classification.
\newblock In {\em International Conference on Neural Information Processing
  (ICONIP)}. Springer, 2017.

\bibitem{phillips2011evaluation}
P.~Jonathon Phillips, Patrick Grother, and Ross Micheals.
\newblock {\em Handbook of Face Recognition}, chapter Evaluation Methods in
  Face Recognition.
\newblock Springer, 2nd edition, 2011.

\bibitem{idiap2012beat}
N~Poh, C~Chan, J~Kittler, J~Fierrez, and J~Galbally.
\newblock Beat--biometrics evaluation and testing: Description of metrics for
  the evaluation of biometric performance.
\newblock Technical report, IDIAP, 2012.

\bibitem{grother2022frvt}
Patrick Grother, Mei Ngan, and Kayee Hanaoka.
\newblock Face recognition vendor test (frvt) part 2: Identification.
\newblock Technical report, National Institute of Standards and Technology,
  2022.

\bibitem{zhang2016joint}
Kaipeng Zhang, Zhanpeng Zhang, Zhifeng Li, and Yu~Qiao.
\newblock Joint face detection and alignment using multitask cascaded
  convolutional networks.
\newblock {\em Signal Processing Letters}, 23(10), 2016.

\bibitem{anjos2012bob}
Andr{\'e} Anjos, Laurent El-Shafey, Roy Wallace, Manuel G{\"u}nther,
  Christopher McCool, and S{\'e}bastien Marcel.
\newblock Bob: a free signal processing and machine learning toolbox for
  researchers.
\newblock In {\em ACM International Conference on Multimedia (ACMMM)}, pages
  1449--1452, 2012.

\bibitem{guenther2012facereclib}
Manuel G{\"u}nther, Roy Wallace, and S{\'e}bastien Marcel.
\newblock An open source framework for standardized comparisons of face
  recognition algorithms.
\newblock In {\em European Conference on Computer Vision (ECCV) Workshops and
  Demonstrations}, pages 547--556. Springer, 2012.

\end{thebibliography}
